\pdfoutput=1

\documentclass[11pt]{article}

\usepackage[preprint]{acl}

\usepackage{times}
\usepackage{latexsym}

\usepackage[T1]{fontenc}

\usepackage[utf8]{inputenc}

\usepackage{microtype}

\usepackage{inconsolata}

\usepackage{graphicx}

\usepackage[utf8]{inputenc} 
\usepackage[T1]{fontenc}    
\usepackage{hyperref}       
\usepackage{url}            
\usepackage{booktabs}       
\usepackage{amsfonts}       
 \usepackage{amsmath}
\usepackage{amssymb}        
\usepackage{nicefrac}       
\usepackage{microtype}      
\usepackage{xcolor}         
\usepackage{graphicx}
\usepackage{amsmath}
\usepackage{adjustbox}
\usepackage{enumitem}
\usepackage{graphicx}
\usepackage{subcaption}
\usepackage{booktabs}
\usepackage{multirow}
\usepackage{amsmath}
\usepackage{float}
\usepackage{pifont}
\usepackage{amssymb}
\usepackage{bbding}
\usepackage{latexsym}
\usepackage{wrapfig}
\usepackage{tablefootnote}
\usepackage{threeparttable}
\usepackage{siunitx}  

\title{Breaking Token Into Concepts: Exploring Extreme Compression in Token Representation Via Compositional Shared Semantics}

\author{Kavin R V \\
  Indian Institute of Technology \\
  Kharagpur, WB, India \\
  \texttt{kavinrv13@gmail.com} \\\And
  Pawan Goyal \\
  Indian Institute of Technology \\
  Kharagpur, WB, India \\
  \texttt{pawang@cse.iitkgp.ac.in} \\}

\begin{document}
\maketitle
\begin{abstract}
Standard language models employ unique, monolithic embeddings for each token, potentially limiting their ability to capture the multifaceted nature of word meanings. We investigate whether tokens can be more effectively represented through a compositional structure that accumulates diverse semantic facets. To explore this, we propose Aggregate Semantic Grouping (ASG), a novel approach leveraging Product Quantization (PQ). We apply ASG to standard transformer architectures (mBERT, XLM-R, mT5) and evaluate this representational scheme across diverse tasks (NLI, NER, QA), as well as a biomedical domain-specific benchmark (BC5CDR) using BioBERT. Our findings demonstrate that representing tokens compositionally via ASG achieves extreme compression in embedding parameters (0.4--0.5\%) while maintaining $>$95\% task performance relative to the base model, even in generative tasks and extends to both cross lingual transfer and domain-specific settings. These results validate the principle that tokens can be effectively modeled as combinations of shared semantic building blocks. ASG offers a simple yet concrete method for achieving this, showcasing how compositional representations can capture linguistic richness while enabling compact yet semantically rich models.

\end{abstract}


\section{Introduction}
\label{sec:introduction}

In modern language models, each token is typically represented by an individual, unique embedding. However, this approach may not be optimal, as semantically similar tokens (e.g., "mother," "mom," and their respective translations in different languages) can be assigned entirely distinct representations, potentially overlooking shared conceptual underpinnings. Recent works \citep{park2023linear, park2024geometry, shani2025tokens} suggests that token representations in LLMs implicitly encode higher-level semantic regularities, often described as concepts, which may be shared across words or subwords. While these studies analyze such concepts as emergent semantic categories or directions in representation space, our work explores an explicit, compositional formulation where tokens are represented as sequences of shared Concept Vectors. In parallel, \newcite{zhang2024tomato} proposed concept-level representations, grouping semantically similar tokens, using k-means. While this method achieved significant vocabulary compression with retained performance, it struggles with polysemy (e.g., "father" as family vs. religious figure) and is limited to encoder-only models, hindered by not explicitly predicting subword in autoregressive decoding.

\begin{figure*} 
    \centering
    \includegraphics[width=\textwidth, keepaspectratio]{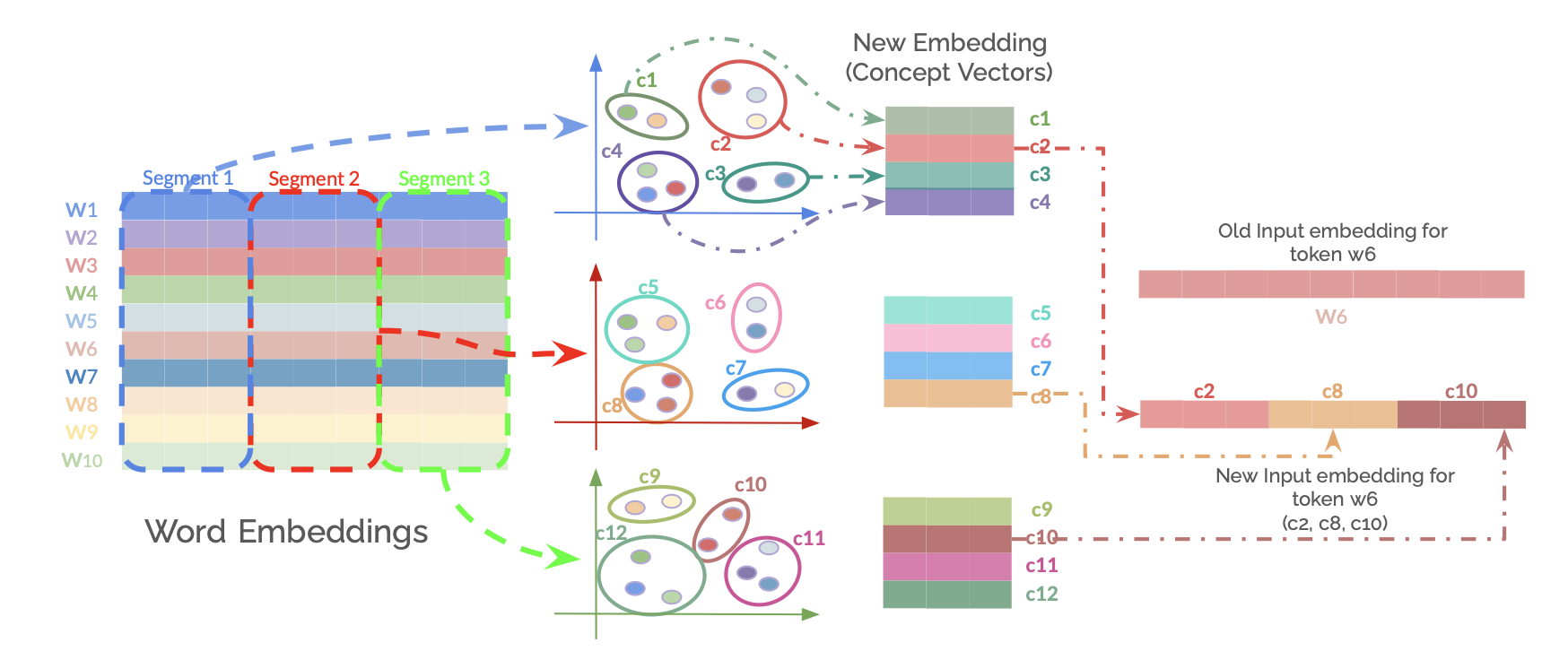}
    \caption{Overview of the Aggregate Semantic Grouping (ASG) method for creating compositional token embeddings. Product Quantization is applied to the original word embedding layer. Embeddings are segmented into $m$ sub-vectors. For each of the $m$ segment positions, k-means clustering is performed on the corresponding sub-vectors from all tokens to learn a codebook of $k$ Concept Vectors (centroids). The new ASG embedding layer containing these learned Concept Vectors is initialized as the embedding layer. Instead of using the original input embedding for a token `w', a sequence of $m$ ConceptIDs used to get their respective Concept Vectors from the ASG layer, these are then concatenated to form the new representation for token `w'. }
    \label{fig:main}
\end{figure*}

To address these limitations, we introduce Aggregate Semantic Grouping (ASG). ASG maintains concept-level sharing but represents tokens as sequences of `conceptIDs', thereby accumulating multiple semantic facets. This sequence-based representation is inspired by successful applications in information and generative retrieval \cite{wang2022neural,tay2022transformer,zhou2022ultron}. We employ Product Quantization (PQ) \citep{5432202} to transform tokens into these conceptID sequences, aiming to preserve token's uniqueness and nuances while benefiting from shared semantics.

Our primary contribution is the introduction of Aggregate Semantic Grouping (ASG), a novel method leveraging Product Quantization to represent tokens as sequences of shared ConceptIDs, thereby capturing multiple semantic facets while significantly compressing embedding layer parameters. We provide a detailed methodology for applying ASG to both encoder and encoder–decoder transformer models. Conducting experiments across diverse tasks (NLI, NER, QA) and models (mBERT, XLM-R, mT5), we demonstrate that even with extreme compression on embeddings (down to 0.4--0.5\% of the original embedding parameters), ASG maintains high performance (often $>$95\% relative to baseline) and outperforms the prior semantic grouping method \citep{zhang2024tomato}, including in zero-shot cross-lingual transfer scenarios. Furthermore, we extend our evaluation to a domain-specific benchmark (BC5CDR; \citealp{DBLP:journals/biodb/LiSJSWLDMWL16}) and a domain-specialized model (BioBERT; \citealp{lee2020biobert}), where ASG achieves similar robustness, confirming its applicability beyond general-domain tasks. The code will be available at \url{https://github.com/KavinRV/Aggregate-Semantic-Grouping}.

\section{Aggregate Semantic Grouping (ASG)}

Our approach, Aggregate Semantic Grouping, reframes token representation by learning compositional embeddings from pre-trained models.

\subsection{Learning Concept Vectors via Product Quantization}
We begin with a pre-trained word embedding matrix $E$, where each row is a $D$-dimensional vector for a token in a vocabulary of size $V$. Using Product Quantization (PQ), each $D$-dimensional embedding is first divided into $m$ distinct segments (sub-vectors), each of dimension $D/m$. For each of these $m$ segment positions, we apply k-means clustering to the collection of all corresponding segments from every token in the vocabulary. This process yields $m$ distinct codebooks; each codebook $C_i$ (for $i=0, \dots, m-1$) contains $k$ centroids, termed \textbf{Concept Vectors}, specific to that segment position. Each Concept Vector is of dimension $D/m$.

\subsection{ASG Embedding Layer Initialization}
The $m$ distinct codebooks ($C_0, C_1, \dots, C_{m-1}$), where each codebook $C_i$ contains $k$ Concept Vectors of dimension $D/m$, are concatenated to form a single, new embedding matrix $E'$. This matrix $E'$ has dimensions $(m \times k) \times (D/m)$ and stores all unique Concept Vectors. Specifically, the $j$-th Concept Vector (where $j \in [0, k-1]$) from the $i$-th codebook $C_i$ is located at row $i \times k + j$ within $E'$.

Each token is then mapped to a sequence of $m$ \textbf{ConceptIDs}. For each of its $m$ embedding segments, the corresponding ConceptID is the specific row index in $E'$ that stores the chosen Concept Vector for that segment. This row index is determined as $i \times k + s_i$, where $i$ is the segment index (from $0$ to $m-1$) and $s_i$ is the index (from $0$ to $k-1$) of the selected centroid from the $i$-th segment's codebook. This sequence of $m$ row indices (ConceptIDs) thus identifies the set of Concept Vectors representing the token.
\begin{table*}[t]
\caption{\label{mbert}
Evaluation results across cluster granularities for \textsc{mBert} and \textsc{XLM-R} on multilingual benchmarks. Scores include F1, Accuracy, and relative performance (\%Base). For XNLI \%Base is for the accuracy relative to the base model. 40\% SG: Semantic Grouping as mentioned in \citet{zhang2024tomato}. In the Zero-Shot setting the models were trained on english dataset and have been tested on all the languages.}
\centering
\resizebox{\textwidth}{!}{
\small
\begin{tabular}{lcc|ccc|cc|cc|cc}
\toprule
\textsc{Model} & \multicolumn{2}{c|}{\textsc{\shortstack{Parameter Reduced \\to (\%)}}} & \multicolumn{3}{c|}{\textsc{XNLI}} & \multicolumn{2}{c|}{\textsc{WikiANN}} & \multicolumn{4}{c|}{\textsc{Zero-Shot}} \\
\cmidrule(lr){9-12}
&&&&&&&&\multicolumn{2}{c|}{\textsc{XNLI}}&\multicolumn{2}{c}{\textsc{WikiANN}}\\
\cmidrule(lr){2-12}
& Embedding & Model & Accuracy & F1 & \%Base & F1 & \%Base & Accuracy & \%Base & F1 & \%Base \\
\midrule
\textsc{mBERT} & 100.00 & 100.00 & 75.46 & 74.79 & 100.00 & 89.74 & 100.00 & 64.86 & 100.00 & 58.58 & 100.00 \\
\textsc{-40\% SG} & 40.00 & 68.95 & 72.43 & 71.88 & 95.99 & 86.69 & 96.61 & 60.64 & 93.49 & 52.35 & 89.37 \\
\textsc{-ASG($k$=512, $m$=48)} & 0.50 & 48.65 & 73.51 & 72.84 & 97.42 & 88.11 & 98.19 & 61.30 & 94.51 & 55.71 & 95.10\\
\midrule
 \textsc{XLM-R} & 100.00 & 100.00 & 77.98 & 77.28 & 100.00 & 88.37 & 100.000 & 71.94 & 100.00 & 58.74 & 100.00 \\
 \textsc{-40\% SG} & 40.00 & 58.48 & 74.56 & 73.96 & 95.61 & 84.57 & 95.70 & 65.83 & 91.51 & 51.48 & 87.65 \\
 \textsc{-ASG($k$=1024, $m$=48)} & 0.40 & 31.08 & 77.06 & 76.39 & 98.81 & 86.53 & 97.92 & 68.05 & 67.39 & 54.46 & 92.72 \\
\bottomrule
\end{tabular}
}
\end{table*}

\subsection{Token Representation with ASG}
When a token is processed, its pre-computed sequence of $m$ ConceptIDs is used to retrieve the corresponding $m$ Concept Vectors from their respective codebooks within $E'$. Let these retrieved Concept Vectors be $v_0, v_1, \dots, v_{m-1}$, where each $v_i$ has dimension $D/m$. The final ASG representation for the token, $e' \in \mathbb{R}^D$, is obtained by concatenating these $m$ Concept Vectors:
\begin{equation}
    e' = \text{concat}(v_0, v_1, \dots, v_{m-1})
    \label{eq:asg_representation_concat}
\end{equation}
This vector $e'$ serves as the input to subsequent layers of the model.

\subsection{Application to Generative Models}
For model with decoder, which have separate input and output embedding layers (the latter often serving as token classifier weights), we apply the ASG process to both. This results in two distinct ASG embedding structures: one for input token representations ($E'$) and another for the output layer ($OE'$), each derived from their respective original embedding matrices.
\begin{table*}[t]
\caption{\label{gst_gen}
Evaluation results for Generative models across cluster granularities for \textsc{mT5} on \textsc{TyDiQA} and \textsc{WikiANN}. Seperate: 1 codebook per segment, Shared: codebooks shared across all segments, In the Zero-Shot setting the models were trained on English dataset and have been tested on all the languages.}
\centering
\resizebox{\textwidth}{!}{
\small
\begin{tabular}{ll|cc|ccc|cc|cc}
\toprule
\multicolumn{2}{c|}{\textsc{Model}} & \multicolumn{2}{c|}{\textsc{\shortstack{Parameter \\ Reduced to (\%)}}} & \multicolumn{3}{c|}{\textsc{TyDIQA}} & \multicolumn{2}{c|}{\textsc{WikiANN}}  & \multicolumn{2}{c}{\textsc{\shortstack{WikiANN \\ (Zero-Shot)}}}\\
\cmidrule(lr){3-4}\cmidrule(lr){5-7} \cmidrule(lr){8-9}\cmidrule(lr){10-11}
 &  & Embedding & Model & F1 & EM & \%Base & F1 & \%Base& F1 & \%Base \\
\midrule
     &\textsc{mT5}                   &  100.00 & 100.00 & 70.74 & 56.20 & 100.00 & 84.21 & 100.00 & 50.75 & 100.00 \\

    \cmidrule(lr){2-11}
 \multirow{4}{*}{\textsc{\shortstack[l]{ASG \\ Separate}}} 
     &\textsc{-($k=1024, m=32$)}
                                        & 0.45 & 15.06 &60.67 & 46.15 & 85.76 & 79.85 & 94.82 & 25.84 & 50.91 \\
 
     &\textsc{-($k=2048, m=32$)}
                                      & 0.85 & 15.41 & 63.81 & 49.06 & 90.19 & 80.93 & 96.11 & 29.85 & 58.82 \\
 
     &\textsc{-($k=8192, m=32$)}
                                      & 3.32 & 17.51 & 66.22 & 51.71 & 93.61 & 82.19 & 97.60 & 33.51 & 66.03 \\
                                      
    &\textsc{-($k=1024, m=64$)}   & 0.45 & 15.06 & 69.96 & 55.53 & 98.89 & 83.18 & 98.78 & 44.02 & 86.74 \\
    \cmidrule(lr){2-11}
 \multirow{3}{*}{\textsc{\shortstack[l]{ASG \\ Shared}}} 
            &\textsc{-($k=16384, m=32$)} & 0.25 & 14.89 & 66.50 & 51.90 & 93.99 & 81.65 & 96.96 & 34.01 & 67.02 \\
            &\textsc{-($k=32768, m=32$)} & 0.45 & 15.06 & 67.00 & 53.06 & 94.71 & 82.04 & 97.42 & 37.01 & 72.92\\
            &\textsc{-($k=32768, m=64$)} & 0.25 & 14.89 & 70.81 & 56.51 & 100.09 & 84.19 & 99.97 & 47.23 & 93.06\\
\midrule
\bottomrule
\end{tabular}
}
\end{table*}

\noindent\textbf{Output Logit Calculation:}
To compute the logit $l_t$ for a target token $t$, the final hidden state $H \in \mathbb{R}^D$ from the model is first segmented into $m$ parts: $H = [H_0, H_1, \dots, H_{m-1}]$, where each $H_i \in \mathbb{R}^{D/m}$. Let the sequence of Concept Vectors for token $t$ be $u_{t,0}, u_{t,1}, \dots, u_{t,m-1} \in OE'$. The logit is calculated as:
\begin{equation}
    l_t = \sum_{i=0}^{m-1} H_i \cdot u_{t,i}
    \label{eq:asg_logit_final}
\end{equation}


\section{Experiments and Results}
\subsection{Datasets}
\label{sec:datasets}
We evaluate our proposed ASG method on diverse cross-lingual benchmarks for natural language inference (NLI), question answering (QA), and named entity recognition (NER). These include: \textbf{XNLI} \citep{conneau2018xnli}, a 15-language sentence understanding benchmark; the Gold Passage (GoldP) task of \textbf{TyDi QA} \citep{clark2020tydi}, an 11-language QA dataset where gold context is provided; and the XTREME benchmark version \citep{hu2020xtreme} of \textbf{WikiANN} \citep{pan-etal-2017-cross}, a 40-language NER dataset.

\subsection{Settings}
For $k$, values were generally chosen as powers of two. This allowed us to systematically target specific levels of embedding parameter compression, aiming for reductions that brought the ASG embedding layer size to approximately 0.5\%, 1\%, and 4\% of the original embedding parameters. Regarding the number of subspaces $m$, our explorations indicated that too few subspaces (e.g., $m=16$) resulted in a significant degradation of model performance. Conversely, using very high values for $m$ (e.g., 128, 256, or 512), would lead to extremely small dimensions for each segment ($D/m$, potentially as low as 4, 2, or 1 for common embedding sizes $D$) and would consequently require very long sequences of ConceptIDs (length $m$) to represent each token. These considerations led us to focus on $m$ values within a moderate range for the experiments detailed below.

\subsection{Fine-tuning Performance}
\label{sec:model_evaluations}
We evaluated ASG on encoder-only mBERT \citep{devlin2019bert} and XLM-R \citep{conneau2019unsupervised} models using the XNLI and WikiANN datasets, mainly to compare it's effectiveness against the Semantic grouping as mention in \citet{zhang2024tomato}. As demonstrated in Table \ref{mbert}, ASG achieves significant embedding compression while maintaining over 97\% of baseline performance and notably outperforms Semantic Grouping (SG) method, even with a low $k$ value. 

To assess ASG for generative tasks, we then evaluated the mT5 model \citep{xue2020mt5} on the TyDiQA and WikiANN datasets, applying ASG to both its input and output embeddings. Table \ref{gst_gen}, detailing results for various cluster ($k$) and subspace ($m$) configurations, shows ASG consistently achieved over 85\% of baseline mT5 performance. Specifically, with $k \ge 2048$, relative performance on TyDiQA surpassed 90\%, while on WikiANN, ASG configurations generally exceeded 95\% of the baseline.

Furthermore, for mT5, we investigated a variant employing a single shared codebook across all $m$ subspaces. To achieve this, the $m$ segments from all token embeddings in the vocabulary are pooled together before applying k-means clustering. This yields one global codebook of Concept Vectors. Each of the $m$ ConceptIDs for a token then selects a Concept Vector from this single shared codebook to represent its corresponding segment. This shared codebook is then used across all $m$ positions for constructing the token representation. This approach, despite reducing the diversity of available Concept Vectors, impressively maintained over 95\% relative performance across both TyDiQA and WikiANN. This suggests that a highly restricted set of output Concept Vectors can still be effective for generative tasks.

\subsection{Domain-Specific Evaluation (BC5CDR)}
\label{sec:bc5cdr}
\begin{table}[ht]
\centering
\small
\begin{tabular}{lcccc}
\toprule
\multirow{2}{*}{\scriptsize\textsc{Model}} & \multicolumn{2}{c}{\scriptsize\textsc{\shortstack{Parameter \\ Reduced to (\%)}}} & \multirow{2}{*}{\scriptsize\textsc{F1}} & \multirow{2}{*}{\scriptsize\textsc{\%Base}} \\
\cmidrule(lr){2-3}
 & \scriptsize\textsc{Emb} &\scriptsize \textsc{Model} & & \\
\midrule
BERT-base & 100 & 100.00 & 85.58 & 100.00 \\
\scriptsize-40\% SG & 40 & 87.64 & 81.57 & 95.32 \\
\scriptsize-ASG ($k{=}128$, $m{=}48$) & 0.81 & 79.50 & 83.90 & 98.03 \\
\scriptsize-ASG ($k{=}512$, $m{=}48$) & 2.13 & 79.77 & 85.24 & 99.60 \\
\midrule
\bottomrule
\end{tabular}
\caption{BERT-base fine-tuned on BC5CDR.}
\label{tab:bc5cdr-bert}
\end{table}
To further validate ASG in specialized settings, we evaluate on the BC5CDR Named Entity Recognition task \cite{DBLP:journals/biodb/LiSJSWLDMWL16}, a biomedical benchmark focused on identifying chemical and disease entities. This task poses strong vocabulary-specific requirements, making it a challenging testbed for compressed embeddings.

We compare BioBERT \cite{lee2020biobert} and BERT-base with standard embeddings, Semantic Grouping (SG), and ASG under varying compression levels. Results are reported in F1 score and relative performance (\%Base).

\begin{table}[ht]
\centering
\small
\begin{tabular}{lcccc}
\toprule
\multirow{2}{*}{\scriptsize\textsc{Model}} & \multicolumn{2}{c}{\scriptsize\textsc{\shortstack{Parameter \\ Reduced to (\%)}}} & \multirow{2}{*}{\scriptsize\textsc{F1}} & \multirow{2}{*}{\scriptsize\textsc{\%Base}} \\
\cmidrule(lr){2-3}
 & \scriptsize\textsc{Emb} &\scriptsize \textsc{Model} & & \\
\midrule
BioBERT & 100 & 100.00 & 89.48 & 100.00 \\
\scriptsize -40\% SG & 40 & 87.64 & 86.71 & 96.90 \\
\scriptsize -ASG($k{=}128$, $m{=}48$) & 0.81 & 79.50 & 87.78 & 98.10 \\
\scriptsize -ASG($k{=}512$, $m{=}48$) & 2.13 & 79.77 & 88.93 & 99.39 \\
\midrule
\bottomrule
\end{tabular}
\caption{BioBERT fine-tuned on BC5CDR.}
\label{tab:bc5cdr-biobert}
\end{table}

Across both backbones, ASG preserves high task performance under strong embedding compression. Even at $<1\%$ of the original embedding size, ASG recovers over 98\% of the base model performance. These results demonstrate that ASG generalizes beyond general-domain benchmarks to biomedical NER.

\subsection{Cross-Lingual Transfer (Zero-Shot)}
\label{sec:cross_lingual_transfer}
For zero-shot cross-lingual transfer, we followed the experimental setup of \newcite{zhang2024tomato}. Models were trained solely on the English XNLI and WikiANN training sets and then evaluated on the multilingual test sets of these datasets. In this setting, ASG-enhanced models outperformed the Semantic Grouping method. While generative models using ASG with lower $k$ (clusters per segment) and $m$ (segments) values showed reduced performance in cross-lingual transfer (\autoref{gst_gen}), configurations with $m=64$ segments nonetheless achieved at least 86\% relative to baseline model performance. Using shared codebook, the performance further improved upto 93\% relative to the baseline model, with just 0.25\% of the embedding parameters.

\subsection{Qualitative Analysis}
\label{app:cluster_examples}
\autoref{fig:ap:father} illustrates how Aggregate Semantic Grouping (ASG) captures varied semantic facets of the token "father" through its clustering across selected segments:

\begin{itemize}[leftmargin=*, itemsep=0pt]
    \item \textbf{Familial Context:} "father" clusters with kinship terms such as "padre" (father), "mother", and "daughter" (Segment 2), or "barn" (child), "parent", and "grandmother" (Segment 12; also Segment 16), reflecting its primary familial sense.
    \item \textbf{Authority/Religious Context:} In Segment 0, "father" groups with "Chief", "Prophet", "notables", and "religión", indicating connotations of leadership or religious reverence.
    \item \textbf{Figurative/Abstract Contexts:} Other segments link "father" to broader concepts, such as "Zeus" (mythological father figure), or with terms like "records", "govern", and "legacy" (Segment 7),  potentially reflecting historical origin, or the act of establishing something significant.
\end{itemize}

\section{Conclusion}
\label{sec:conclusion}

This work investigated equipping language models with shared, compositional token representations as an alternative to traditional monolithic embeddings. We explored this through Aggregate Semantic Grouping (ASG), where Product Quantization transforms embeddings into sequences of \textbf{ConceptIDs} that map to shared, learned \textbf{Concept Vectors}, enabling multifaceted semantic capture alongside significant compression. Extensive experiments on diverse models (including mBERT, XLM-R, and mT5) and NLU tasks (such as NLI, NER, and QA) found ASG maintains high performance (often >95\% relative to baseline) despite extreme parameter reduction (to <1\% of original size). ASG also outperformed prior semantic grouping methods, and proved effective for generative architectures. These findings confirm that ASG's decomposition of tokens into shared components offers an efficient, semantically rich, and promising direction for language modeling; future work may explore dynamic or adaptive quantization techniques.

\section{Limitations}
\label{sec:limitations}

ASG was applied directly to word embeddings from pre-trained models without an explicit cross-lingual alignment step, which could refine Concept Vector clustering. This may partly explain the observed performance degradation in generative tasks within cross-lingual settings, such as on WikiANN, where better nuance preservation through alignment-optimized clustering could be beneficial. Furthermore, we did not undertake continual pre-training of the models with the ASG embeddings; such a phase could allow models to more effectively adapt to the compositional representations and potentially enhance overall performance. Complementary to this, methods similar GraphMerge \cite{wu-monz-2023-beyond}, could potentially be combined with ASG to pre-align embeddings before clustering, leading to more coherent ConceptID assignments.

\bibliography{custom}

\clearpage

\appendix
\begin{table*}[ht]
\centering
\caption{Model Configurations and Embedding Parameters for ASG, Underlined uses a shared codebook}
\label{tab:asg_model_configurations}
\begin{tabular}{l c c c c}
\toprule
\textbf{Model} & \textbf{k} & \textbf{m} & \textbf{Parameters} & \textbf{Embedding Shape (Dim)} \\
\midrule
\multirow{2}{*}{mBERT} & N/A & N/A & ~177M & [~120k, 768] \\
                       & 512 & 48  & ~86M  & [~30k, 16] \\
\midrule
\multirow{2}{*}{XLM-R} & N/A & N/A & ~277M & [~250k, 768] \\
                       & 1024& 48  & ~86M   & [~49k, 16]\\
\midrule
\multirow{8}{*}{mT5} & N/A & N/A  & ~300M & [~256k, 512] \\
                       & 1024& 32 & ~45M  & [~36k, 16]   \\
                       & 2048& 32 & ~46M  & [~68k, 16]   \\
                       & 8192& 32 & ~53M  & [~265k, 16]  \\
                       & \underline{16384} & 32 & ~44M & [~20k, 16] \\
                       & \underline{32768} & 32 & ~45M & [~36k, 16] \\
                       & \underline{32768} & 64 & ~44M & [~39k, 8] \\
                       & 1024& 64 & ~45M & [~72k, 8] \\ 
\bottomrule
\end{tabular}
\end{table*}

\section{Experimental Setup}
\label{sec:experimental_setup}

All our experiments are conducted using the smallest available checkpoint for each respective pre-trained model. Training is performed with a batch size of 128, and all experiments were run on a single Nvidia L40 GPU.

For the encoder models (mBERT and XLM-R), we set a weight decay of $0.01$. The learning rate was $5 \times 10^{-6}$ for XNLI experiments and $5 \times 10^{-5}$ for WikiANN experiments. These models were trained for 2 epochs; for cross-lingual transfer settings, training was extended to 5 epochs. The mT5 model was trained with a learning rate of $1 \times 10^{-3}$.

Product Quantization is implemented using the \texttt{nanopq} library\footnote{\url{https://github.com/matsui528/nanopq}}. For the k-means clustering within \texttt{nanopq} we use the \texttt{faiss} library\footnote{\url{https://faiss.ai/}}.

\section{Parameter Reduction Calculation}
\label{appendix:paramclarification}
In a standard model, the token embedding table has shape $(V, D)$, where $V$ is the vocabulary size and $D$ is the embedding dimension.  

With ASG, the embedding layer is replaced by $m$ codebooks, each with $k$ Concept Vectors of size $D/m$. The ASG embedding matrix thus has shape:
\[
(k \cdot m, \tfrac{D}{m})
\]

The number of parameters becomes:
\[
k \cdot m \cdot \tfrac{D}{m} = k \cdot D
\]

So the ratio of ASG to original embedding parameters is:
\[
\frac{k \cdot D}{V \cdot D} = \frac{k}{V}
\]

For example, in XLM-R with $V=250{,}000$, $k=1024$, and $m=48$, the ASG matrix has shape $(49{,}000, 16)$, and the compression ratio is:
\[
\frac{49{,}000 \cdot 16}{250{,}000 \cdot 768} \approx 0.0040 \, (=0.4\%).
\]

\section{Token-to-ConceptID Mapping Matrix.}
Each token is mapped to a sequence of $m$ ConceptIDs (integers in $[0,k)$), forming a matrix of shape $(V, m)$. This mapping is:
\begin{itemize}
    \item Not a parameter: it is precomputed, fixed, and non-trainable.
    \item Stored externally and could be used during tokenization.
    \item Extremely compact, as it contains only small integers.
\end{itemize}

For $V=250$k and $m=48$, this mapping can be stored with an overhead of $\sim 2\%$ ($15$MB) or $\sim 2.15\%$ ($16.5$MB) of the memory required for the original embedding matrix, when $k=1024$ or $k=2048$ respectively.  

\section{ASG Configuration and Embedding Parameters}
\label{app:asg_config_params}

Table~\ref{tab:asg_model_configurations} provides a summary of the configurations used for Aggregate Semantic Grouping (ASG) across different models, alongside details for the original base models. The table specifies the choices for $k$ (number of centroids per subspace) and $m$ (number of subspaces) for each ASG setup. It also lists the resulting total number of parameters in the model and the shape of the embbeding layer.





\begin{figure*}[ht]
    \includegraphics[width=\textwidth]{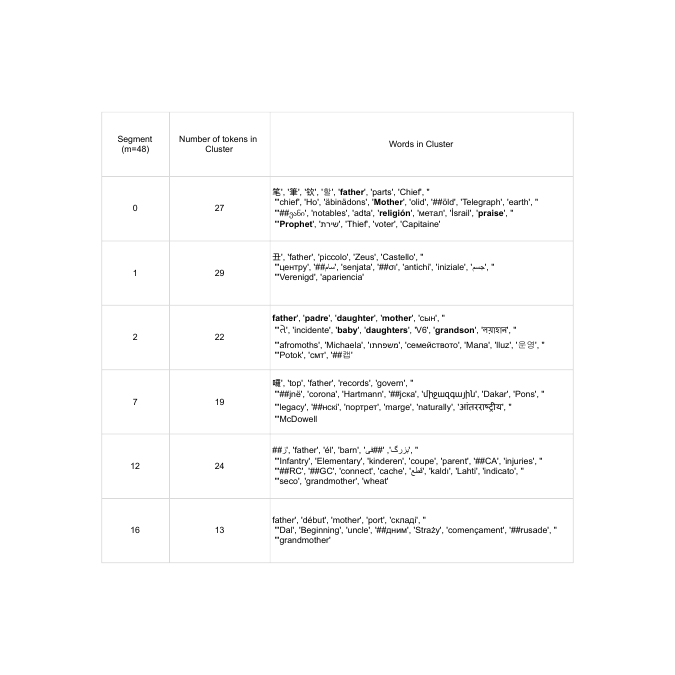} 
    \caption{Grouping of the token "father" at a few selected subspaces}
    \label{fig:ap:father}
\end{figure*}

\end{document}